\documentclass{article}
\usepackage{ijcai19}
\usepackage{amsmath}
\usepackage{amsfonts}
\usepackage{amssymb}
\usepackage{times}
\usepackage{latexsym}
\usepackage[utf8]{inputenc}
\usepackage[T1]{fontenc}

\usepackage{booktabs}
\usepackage{graphicx}
\usepackage{xcolor}
\usepackage{comment}
\def\mc#1#2{\multicolumn{#1}{c}{#2}}
\def\mr[#1]#2#3{\multirowcell{#2}[#1]{#3}}
\usepackage{array,multirow,graphicx}

\title{ Attention-based Modeling for Emotion Detection and Classification in Textual Conversations\footnote{This work was presented at 2nd Workshop on Humanizing AI (HAI) at IJCAI'19 in Macao, China.}}
\author{
Waleed Ragheb$^{1,2}$ \and 
 Jérôme Azé $^{1,2}$ \and
Sandra Bringay$^{1,3}$ \and
Maximilien Servajean$^{1,3}$
\affiliations
$^1$LIRMM UMR 5506, CNRS, University of Montpellier, Montpellier, France\\
$^2$IUT de Béziers, University of Montpellier, Béziers, France\\
$^3$AMIS, Paul  Valery University - Montpellier 3 , Montpellier, France
\emails
\{First.Last\}@lirmm.fr
}

\begin{document}
\maketitle

\begin{abstract}
This paper addresses the problem of  modeling textual conversations and  detecting emotions. Our proposed model makes use of 1) deep transfer learning rather than the classical shallow methods of word embedding; 2)  self-attention mechanisms to focus on the most important parts of the texts and 3) turn-based conversational modeling for classifying the emotions. The approach does not rely on any hand-crafted features or lexicons.
Our model was evaluated on the data provided by the SemEval-2019 shared task on contextual emotion detection in text. The model shows very competitive results.

\end{abstract}

\section{Introduction}
Emotional intelligence has played a significant role in many application in recent years \cite{Krakovsky:2018:AI:3200906.3185521}.
It is one of the essential abilities to move from narrow to general human-like intelligence. Being able to recognize expressions of human emotion such as interest, distress, and pleasure in communication is vital for helping machines choose more helpful and less aggravating behavior. Human emotions are a mental state that can be sensed and hence recognized in many sources such as visual features in images or videos \cite{emoimg}, as textual semantics and sentiments in texts \cite{emotext} or even patterns in  EEG brain signals \cite{egg}. With the increasing number of messaging platforms and with the growing demand of customer chat bot applications, detecting the emotional state in conversations becomes highly important for more personalized and human-like conversations \cite{5-chatbot}.

This paper addresses the problem of modeling a conversation that comes with multiple turns for detecting and classifying emotions. The proposed model makes use of transfer learning through the universal language modeling that is composed of consecutive layers of Bi-directional Long Term Short Term Memory (Bi-LSTM) units. These layers are learned first in sequence-to-sequence fashion on a general text and then fine-tuned to a specific target task. The model also makes use of an attention mechanism in order to focus on the most important parts of each text turn. Finally, the proposed classifier models the changing of the emotional state of a specific user across turns.% of conversations. 

This article is an extension of the work done for the Semeval-2019 Task-3  \cite{SemEval2019Task3} including a discussion on the identification of vocabulary related to feeling by the attention layer. The  paper is organized as follows. In Section \ref{RW}, the related work is introduced. Then, we present a quick overview of the task and the datasets in Section \ref{DS}. Section \ref{PMS} describes the proposed model architecture, some variants and hyperparameters settings. The experiments and results are presented in Section \ref{XPR}. Section \ref{CN} concludes the study.

\section{Related Work}\label{RW}

Transfer learning or domain adaptation has been widely used in machine learning especially in the era of deep neural networks \cite{Goodfellow-et-al-2016}. 
\begin{comment}
It enables reusing the models developed and trained in a source task to another target task. The power of transfer learning is very clear when the features learned from the source or base task are general and can be repurpose to the target tasks. Computer Vision (CV) models are the most common and widely used models that make use of domain adaptation. Nowadays, most of CV models base extracting the feature to a pretrained models like AlexNet, ResNet, MS-COCO ... etc \cite{cvPre}. 
\end{comment}
In natural language processing (NLP), this is done through Language Modeling (LM). Through this step, the model aims to predict a word given some context. This is considered as a vital and important basics in most of NLP applications. Not only because it tries to understand the long-term dependencies and hierarchical structure of the text but also for its open and free resources. LM is considered as unsupervised learning process which needs only corpus of unlabeled text. The problem is that LMs get overfitted to small datasets and suffer catastrophic forgetting when fine-tuned with a classifier. Compared to Computer Vision (CV), NLP models are typically more shallow and thus require different fine-tuning methods. The developing of the Universal Language Model Fine-tuning (ULMFiT) \cite{jerm} is considered like moving from shallow to deep pre-training word representation. This idea has been proved to achieve CV-like transfer learning for many NLP tasks. ULMFiT makes use of  the state-of-the-art AWD-LSTM (Average stochastic gradient descent - Weighted Dropout) language model \cite{meri_1}. Weight-dropped LSTM is a strategy that uses a DropConnect \cite{DropConnect} mask on the hidden-to-hidden weight matrices, as a means to prevent overfitting across the recurrent connections. %The same 3-layer LSTM architecture with the same hyperparameters are used. The overall model with a simple avegare and maximum pooling classifier signiﬁcantly outperforms the state-of-the-art on six text classification tasks including three tasks for sentiment analysis. 

On the other hand, one of the recent trend in deep learning models is the attention Mechanism \cite{recentNLP}. Attention in neural networks are inspired from the visual attention mechanism found in humans. The main principle is being able to focus on a certain region of an image with “high resolution” while perceiving the surrounding image in “low resolution”, and then adjusting the focal point over time. This is why the early applications for attention were in the field of image recognition and computer vision \cite{NIPS2010_4089}. In NLP, most competitive neural sequence transduction models have an encoder-decoder structure \cite{att_all}. A limitation of these architectures is that it encodes the input sequence to a fixed length internal representation. This cause the results going worse performance for very long input sequences. Simply, attention tries to overcome this limitation by guiding the network to learn where to pay close attention in the input sequence. Neural Machine Translation (NMT) is one of the early birds that make use of attention mechanism \cite{nmt}. It has recently been applied to other problems like sentiment analysis \cite{aspectS} and emotional classification \cite{emoOrg}.

\section{Data}\label{DS}
The datasets provided by the task organizers of Semeval-2019 Task-3 are collections of labeled conversations \cite{SemEval2019Task3}. Each conversation is a three turn talk between two persons. The conversation labels correspond to the emotional state of the last turn. Conversations are manually classified into three emotional states for \textit{happy}, \textit{sad}, \textit{angry} and one additional class for \textit{others}.  In general, released datasets are highly imbalanced and contains about 4\% for each emotion in the validation (development) set and final test set. Table \ref{DT} shows the number of conversations examples and emotions provided in the official released datasets.

\begin{table}[htbp]
\centering % used for centering table
%\captionsetup{justification=centering}
 % title of Table

\resizebox{\columnwidth}{!}{\begin{tabular}{lrrrr}% centered columns (4 columns)
\toprule %inserts double horizontal lines
 Dataset &  Data size & Happy & Sad & Angry \\  % inserts table
%heading
\midrule % inserts single horizontal line
Training & 30160 & 5191 & 6357 & 6027\\ % inserting body of the table
Validation (Dev) & 2755 & 180 & 151 & 182\\  % [1ex] adds vertical space
Testing &  5509 & 369 & 308 & 324\\
\bottomrule %inserts single line
\end{tabular}}
\caption{Used datasets.}
\label{DT}
\end{table}

\section{Proposed Models}\label{PMS}
In this section, we present the proposed model architecture for modeling a conversation through language models encoding and classification stages. Also, we explain the the training procedures used and the external resources for training the language model. In addition to the basic architecture, We will describe the used variants of the model for evaluation. Finally, we will list the hyperparameters used for building and training these models. 

\subsection{Model Architecture}
\begin{figure*}
\centering
\def\svgwidth{\columnwidth}
\includegraphics [width=\linewidth]{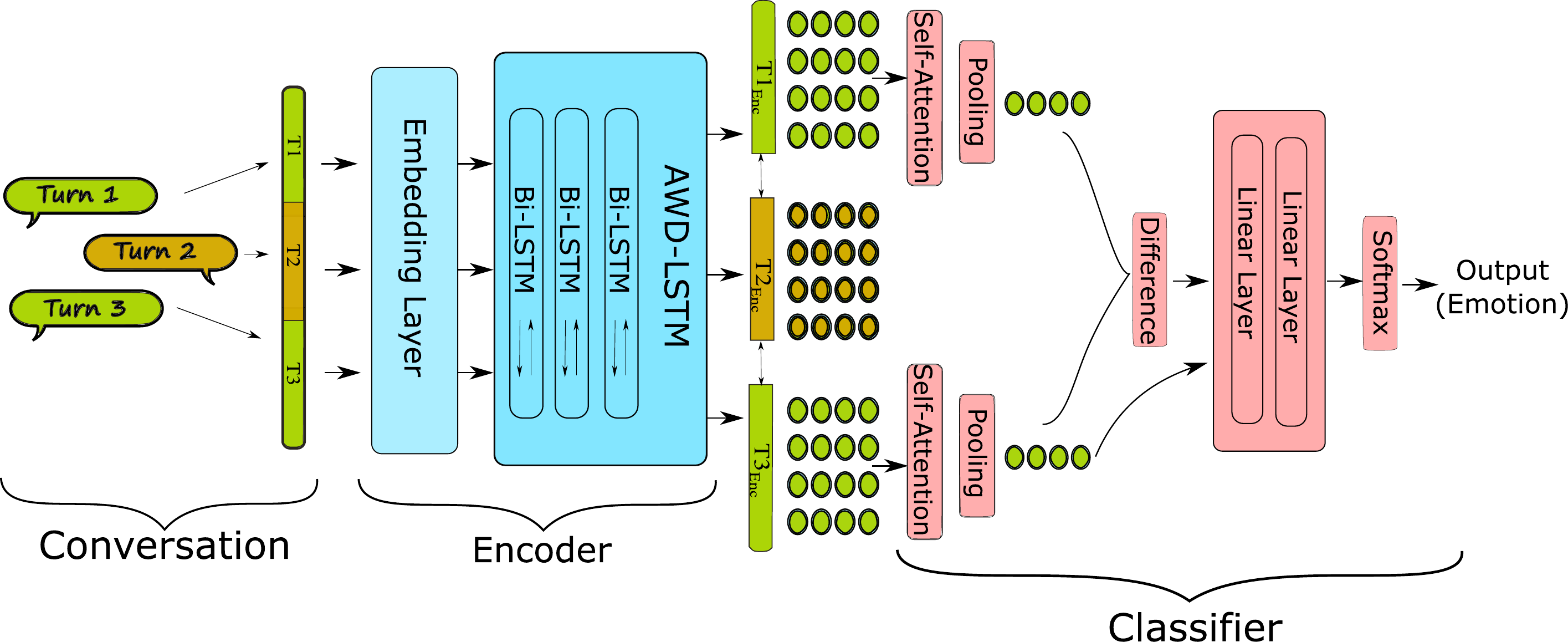}
\caption{Proposed model architecture (\textit{Model-A}).}
\label{pma}
\end{figure*}
In figure \ref{pma}, we present our proposed model architecture. The model consists of two main steps: encoder and classifier. We used a linear decoder to learn the language model encoder as we will discuss later.  This decoder is replaced by the classifier layers. The input conversations come in turns of three. After tokenization, we concatenate the conversation text but keep track of each turn boundaries. The overall conversation is inputted to the encoder. The encoder is a normal embedding layer followed by AWD-LSTM block. This uses three stacked different size Bi-LSTM units trained by ASGD (Average Stochastic Gradient Descent) and managed dropout between LSTM units to prevent overfitting. The conversation encoded output has the form of $C_{Enc}=[T_{Enc}^1 \oplus T_{Enc}^2 \oplus T_{Enc}^3]$
where $T^i$ is the i\textsuperscript{th} turn in the conversation and $\oplus$ denotes a concatenation operation and $T_{Enc}^i = \{T_1^i,T_2^i, \ldots, T_{N_i}^i\}$. The sequence length of turn $i$ is denoted by $N_i$. The size of $T_j^i$ is the final encoding of the $j$'s sequence item of turn $i$.

For classification, the proposed model pays close attention to the first and last turns. The reasons behind this are that the problem is to classify the emotion of the last turn. Also, the effect of the middle turn appear implicitly on the encoding of the last turn as we used Bi-LSTM encoding on the concatenated conversation. In addition to these, tracking the difference between the first and the last turn of the same person may be beneficial in modeling the semantic and emotional changes. So, we apply self-attention mechanism followed by an average pooling to get turn-based representation of the conversation. The attention scores for the i\textsuperscript{th} turn $S^i$ is given by:

\begin{equation}
    S^i=Softmax \{W_i.T_{Enc}^i\}
\end{equation}

Where $W_i$ is the weight of the attention layer of the $i^{th}$ turn and $S^i$ has the form of $S^i = \{S_1^i,S_2^i, ..., S_{N_i}^i\}$. The output of the attention layer is the scoring of the encoded turn sequence $O^i=\{o_1^i,o_2^i,\dots ,o_{N_i}^i\}$ which has the same length as the turn sequence and is given by $O^i = S^i \odot T_{Enc}^i$
where $\odot$ is the element-wise multiplication. The difference of the pooled scored output of $O^1$ and $O^3$ is computed as $O_{\text{diff}}$. The Input of the linear block is $X_{\text{in}}$ is formed by:

\begin{equation}
    X_{in} = [O_{\text{diff}} \oplus O_{\text{pool}}^3]
\end{equation}

The fully connected linear block consist of two different sized dense layers followed by a Softmax to determine the target emotion of the conversation. 

\subsection{Training Procedures}
Training the overall models comes into three main steps:
\begin{enumerate}
\item The LM is randomly initialized and then trained by stacking a linear decoder in top of the encoder. The LM is trained on a general-domain corpus. This helps the model to get the general features of the language.
\item The same full LM after training is used as an initialization to be fine-tuned using the data of the target task (conversation text). In this step we limit the vocabulary of the LM to the frequent words (repeated more tan twice) of target task.
\item We keep the encoder and replace the decoder with the classifier and both are fine-tuned on the target task.% using different strategies for each layer group.
\end{enumerate}

For training the language model, we used the Wikitext-103 dataset \cite{wiki}.
\begin{comment}With more than 28K of Wikipedia articles and 103 million words, the model determine the main structure and hierarchy of the language by sequence-to-sequence modeling. The training of the architecture is done using  slanted triangular learning rates (STLR) which change the learning rate for each iteration in triangular fashion. we used only once cycle as recommend in \cite{jerm}. The model was trained by discriminative fine-tuning which uses different learning rate for each layer group. The model is trained gradually by freezing and unfreezing layers for different groups. 
\end{comment}
We train the model on the forward and backward LMs for both the general-domain and task specific datasets. Both LMs -backward and forward- are used to build two versions of the same proposed architecture. The final decision is the ensemble of both.  Our code is released at https://github.com/WaleedRagheb/Attentive- Emocontext. However we tried the uni-directional models, experimental studies shows that the ensemble models give a better performance. Training the self-attention layer uses the same learning rates used in the classification layers group.

 We used Pytorch \footnote{https://pytorch.org/} to build the whole model and make use of Fastai  \footnote{http://www.fast.ai/} libraries for applying the training strategies and fine-tuning the language models. For text preprocessing, the text is first normalized and tokenized. Special tokens were added for capitalized and repeated words. we keep the punctuation and the emotions symbols in text. We used Spacy \footnote{https://spacy.io/} and the wrapper of FastText \footnote{https://fasttext.cc/}. The models are trained and tested on Nvidia GEFORCE GTX 1080 GPU.

\subsection{Model Variations}
In addition to the model  \textit{- Model-A -}   described by Figure \ref{pma}, we tried five different variants. %different variants of the same architecture. 
Each variant modify the classifier layer groups. Studying the effect of these variants will provide a good model ablation analysis. 
%We will refer to the basic model  as . 

The first variant -(\textit{Model-B})- is formed by bypassing the self attention layer. This will pass the output of the encoder directly to the average pooling layer such that $X_{\text{in}}^B = [T_{\text{diff}} \oplus T_{\text{pool}}^3]$ where $T_{\text{diff}}$ is the difference between the first and third pooled encoded turns of the conversations. 

-(\textit{Model-C})- is to input a pooled condensed representation to the whole conversation $C_{\text{pool}}$ rather than the last turn to the linear layer block. In this case: $X_{\text{in}}^C = [O_{\text{diff}} \oplus C_{\text{pool}}]$.
We also studied two versions of the basic model where only one input is used %- either the turn difference or the representation of last turn;
$X_{\text{in}}^D = O_{\text{diff}}$ -(\textit{Model-D})- and $X_{\text{in}}^E = O_{\text{pool}}^3$ -(\textit{Model-E}).
In these two variants, we just change the size of the first linear layer. 

Also, we apply the forward direction LM and classifier only without ensemble them with the backward direction and keep the same basic architecture -(\textit{Model-F}).

\begin{table*}[t]
\setlength\tabcolsep{5pt}
\centering
  \resizebox{\linewidth}{!}{\begin{tabular}{*{11}{lrrrr}}
   \toprule
      
    & \mc{10}{Results} \\
   \cmidrule{2-11}
\mc{1}{Models}   & \mc{3}{Happy}
    & \mc{3}{Sad}
    & \mc{3}{Angry}
    & \mc{1}{Micro}
     \\
      \cmidrule(lll){2-4}
      \cmidrule(lll){5-7}
      \cmidrule(lll){8-10}
      \cmidrule(l){11-11}
      
    & \multicolumn{1}{c}{P} & \multicolumn{1}{c}{R} & \multicolumn{1}{c}{F1} & \multicolumn{1}{c}{P} & \multicolumn{1}{c}{R} & 
    \multicolumn{1}{c}{F1} & \multicolumn{1}{c}{P} & \multicolumn{1}{c}{R} & \multicolumn{1}{c}{F1}& \multicolumn{1}{c}{F1}\\
   \midrule
    \multicolumn{1}{c}{A} & 0.7256&0.7077&\textbf{0.7166}&0.8291&0.776&\textbf{0.8017}&0.7229&0.8054&\textbf{0.7619}&\textbf{0.7582} \\
     \midrule
    \multicolumn{1}{c}{B} & \textbf{0.7341}&0.6514&0.6903&0.7401&\textbf{0.82}&0.778&0.7049&\textbf{0.8255}&0.7604&0.7439 \\
    \multicolumn{1}{c}{C} & 0.7279&0.6972&0.7122&0.7765&0.792&0.7842&0.6941&0.8221&0.7527&0.7488 \\
    \multicolumn{1}{c}{D} & 0.7214&\textbf{0.7113}&0.7163&0.8128&0.764&0.7876&0.6965&0.8087&0.7484&0.749 \\
    \multicolumn{1}{c}{E} & 0.7204&0.7077&0.714&0.8205&0.768&0.7934&0.7026&0.8087&0.752&0.7512 \\
    \multicolumn{1}{c}{F} & 0.7336&0.669&0.6998&\textbf{0.8377}&0.764&0.7992&\textbf{0.738}&0.7752&0.7561&0.75 \\
  
   \bottomrule
  \end{tabular}}
  \caption{Test set results of the basic proposed model and its variants.}
  \label{Res_t}
\end{table*}

\subsection{Hyperparameters}
We use the same set of hyperparameters across all model variants. For training and fine-tuning the LM, we use the same set of hyperparameter of AWD-LSTM proposed by \cite{meri_1} replacing the LSTM with Bi-LSTM and keep the same embedding size of $400$ and $1150$ hidden activations. We used weighted dropout of $0.2$ and $0.25$ as the input embedding dropout and the learning rate is $0.004$. We fine-tuned the LM by all provided datasets in table \ref{DT}. We train the LM for $14$ epochs using batch size of $128$ and limit the number of vocabulary to all token that appear more than twice. 
 For classifier, we used masked self-attention layers and average pooling. For the linear block, we used hidden linear layer of size $100$ and apply dropout of $0.4$. We used Adam optimizer \cite{adam} with $\beta_1=0.8$ and $\beta_2=0.99$. The base learning rate is $0.01$. We used the same batch size used in training LMs but we create each batch using weight random sampling. We used the same weights provided by the organizers 
(0.4 for each emotion). We train the classifier on training set for 30 epochs and select the best model on validation set to get the final model.

\section{Results \& Discussions}\label{XPR}
The results of the test set for different variants of the model for each emotion is shown in table \ref{Res_t}. The table shows the value of precision (P), recall (R) and F1 measure for each emotion and the micro-F1 for all three emotional classes. The micro-F1 scores are the official metrics used in this task. \textit{Model-A} gives the best performance F1 for each emotion and the overall micro-F1 score. However some variants of this model give better recall or precision values for different emotions, \textit{Model-A} compromise between these values to give the best F1 for each emotion. Removing the self-attention layer in the classifier -\textit{Model-B}- degraded the results. Also, inputting a condensed representation of the all conversation rather than the last turn -\textit{Model-C}- did not improve the results. Even modeling the turns difference only -\textit{Model-D}- gives better results over \textit{Model-C}. These proves empirically the importance of the last turn in the classification performance. This is clear for \textit{Model-E} where the classifier is learned only by inputting the last turn of the conversation. Ensemble the forward and backward models was more useful than using the forward model only -\textit{Model-F}.

Comparing the results for different emotions and different models, we notice the low performance in detecting happy emotion. This validate the same conclusion of Chatterjee et.al in \shortcite{CHATTERJEE2019309}. They justify this by the difficulties even for human level annotation to discriminate between happy and many other emotions. 
The model shows a significant improvement over the EmoContext organizer baseline (F1: $0.5868$). Also, comparing to other participants in the same task with the same datasets, the proposed model gives competitive performance and ranked 11\textsuperscript{th} out of more than 150 participants. The proposed model can be used to model multi-turn and multi-parties conversations. It can be used also to track the emotional changes in long conversations.

One of the most attractive outcomes of applying the attention mechanism is its ability to process all the input sequences with different weights of attentions. It usually pays closer attention to the most important parts that influence the network decision. To validate these findings, we compared the most important tokens in terms of attention scores with sentiment and emotional lexicons. We found EmoLex proposed by S.Mohammad et al. in \shortcite{seif_2} \shortcite{seif_1} a good example. EmoLex is created with a high-quality, moderate-sized, emotion and polarity lexicon. It has entries for more than 10,000 word-sense pairs. We extracted the words related Emocontext emotions for happy (Joy) and sad (sadness) and angry (anger). Table \ref{Res_at} shows the results of matching the top 20\% attention scored tokens with EmoLex in both validation (Dev) and testing testsets. The self-attention layers proposed in the first $T_1$ and last turn $T_3$ in the conversation seems to pay close attention to the corresponding emotional words. This is clear with the diagonal in table \ref{Res_t}. However the mentioned difficulties in Happy emotion detection, the self-attention focuses in parts of text related to joy with a significant difference between the sadness and anger lexicon words. This significance is decreased between the sadness and anger words. However, the attention model is well focused to the correct emotions.

\begin{table}[h]
\centering
\begin{tabular}{cccccc}
\toprule
%\multicolumn{1}{|c|}{label01} & \multicolumn{3}{c|}{label02}      
&&\multicolumn{3}{c}{\underline{Lexicon-based}}\\
%\cmidrule{3-5}
&Datasets&\multicolumn{1}{c}{Joy} & \multicolumn{1}{c}{Sadness}& \multicolumn{1}{c}{Anger}\\
\cmidrule{3-5}
\parbox[b]{2mm}
{\multirow{7}{*}{\centering \rotatebox[origin=c]{90}{\underline{Attention-based}}}}
\multirow{2}{*}{  Happy}
&(V)& \textbf{42.57\%} & 4.95\% & 4.05\% \\ %\cline{3-4} 
                                                                            & (T) & \textbf{39.27\%} & 7.97\% & 7.36\% \\ \cmidrule{2-5} 
                              \multirow{2}{*}{Sad}              &    (V)     & 21.66\% & \textbf{40.58\%} & 23.04\% \\ %\cline{3-4} 
                                                                            & (T) & 20.59\% & \textbf{32.25\%} & 26.04\%  \\
                             \cmidrule{2-5} 
                                 \multirow{2}{*}{     Angry}                  & (V)     & 21.07\% & 26.05\% & \textbf{39.73\%} \\ %\cline{3-4} 
                                                                  & (T)          & 22.02\% & 22.97\%  & \textbf{35.02\%}\\
                             
\bottomrule

\end{tabular}
\caption{Matching Percentages of emotion related words in the top 20\% attention scored parts of the text in $T_1$ and $T_2$ in Validation (V) and Testing (T) datasets .}
  \label{Res_at}
\end{table}

\section{Conclusions}\label{CN}
In this paper, we present a new model used for Semeval-2019 Task-3 \cite{SemEval2019Task3}. The proposed model makes use of deep transfer learning rather than the shallow models for language modeling. The model pays close attention to the first and the last turns written by the same person in 3-turn conversations. The classifier uses self-attention layers and the overall model does not use any special emotional lexicons or feature engineering steps. The results of the model and its variants show a competitive results compared to the organizers baseline and other participants. Our best model gives micro-F1 score of 0.7582. The model can be applied to other emotional and sentiment classification problems and can be modified to accept external attention signals and emotional specific word embedding. 

\section*{Acknowledgement}
We would like to acknowledge La Région Occitanie and Communauté d'Agglomération Béziers Méditerranée which finance the thesis of Waleed Ragheb as well as  INSERM and CNRS for their financial support of CONTROV project.

\bibliographystyle{named}
\bibliography{Lirmm_Emocontext}

\end{document}